\documentclass[11pt,a4paper]{article}
\usepackage[hyperref]{acl2020}
\usepackage{times}
\usepackage{latexsym}

\usepackage{microtype}

\usepackage{enumitem}
\usepackage{amsmath}
\usepackage{graphicx}

\aclfinalcopy 

\renewcommand{\sectionautorefname}{\S\kern-0.2em}
\renewcommand{\subsectionautorefname}{\S\kern-0.2em}

\newcommand{\modelname}[1]{ \textsc{ModelName}}

\title{A Recipe for Creating Multimodal\\
 Aligned Datasets for Sequential Tasks}

\newcommand{\affilmsr}{${}^{\diamondsuit}$}
\newcommand{\affilsf}{${}^{\clubsuit}$}
\author{Angela S. Lin \affilsf\thanks{\hspace{0.2em} Work done when the author was an intern at Microsoft.}  \quad Sudha Rao\affilmsr \quad Asli Celikyilmaz\affilmsr \quad Elnaz Nouri\affilmsr \quad \\
  \bf Chris Brockett\affilmsr \quad Debadeepta Dey\affilmsr \quad Bill Dolan\affilmsr \\
  \affilsf Salesforce Research, Palo Alto, CA, USA \\
  \affilmsr Microsoft Research, Redmond, WA, USA \\
  {\tt \footnotesize angela.lin@salesforce.com \{sudhra, aslicel, elnouri\}@microsoft.com} \\
  {\tt \footnotesize \{chrisbkt, dedey, billdol\}@microsoft.com}}

\date{}

\begin{document}
\maketitle
\begin{abstract}

Many high-level procedural tasks can be decomposed into sequences of instructions that vary in their order and choice of tools. 
In the cooking  domain, the web offers many partially-overlapping text and video recipes (i.e. procedures) that describe how to make the same dish (i.e. high-level task). 
Aligning instructions for the same dish across different sources can yield descriptive visual explanations that are far richer semantically than conventional textual instructions, providing commonsense insight into how real-world procedures are structured. 
Learning to align these different instruction sets is challenging because: a) different recipes vary in their order of instructions and use of ingredients; and b) video instructions can be noisy and tend to contain far more information than text instructions.
To address these challenges, we first use an unsupervised alignment algorithm that learns pairwise alignments between instructions of different recipes for the same dish.
We then use a graph algorithm to derive a joint alignment between multiple text and multiple video recipes for the same dish.
We release the \textsc{Microsoft Research Multimodal Aligned Recipe Corpus}\footnote{\url{https://github.com/microsoft/multimodal-aligned-recipe-corpus}} containing $\sim$150K pairwise alignments between recipes across 4,262 dishes with rich commonsense information. 

\end{abstract}

\section{Introduction}

\begin{figure}
    \centering
    \includegraphics[trim=95 120 0 100, scale=0.35]{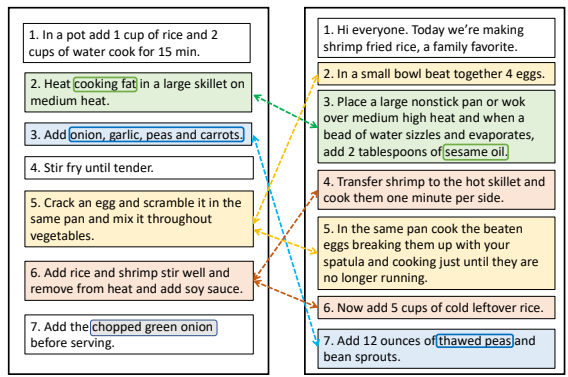}
    \caption{Text recipe (left) and transcript of video recipe (right) for \textit{\textbf{shrimp fried rice}}. Aligned instructions are highlighted in the same color. Ingredients that can be substituted are encircled in the same color. }
    \label{fig:fried_rice_recipe_pair}
\end{figure}

\begin{figure*}
    \centering
    \includegraphics[trim=0 120 0 100, width=\textwidth]{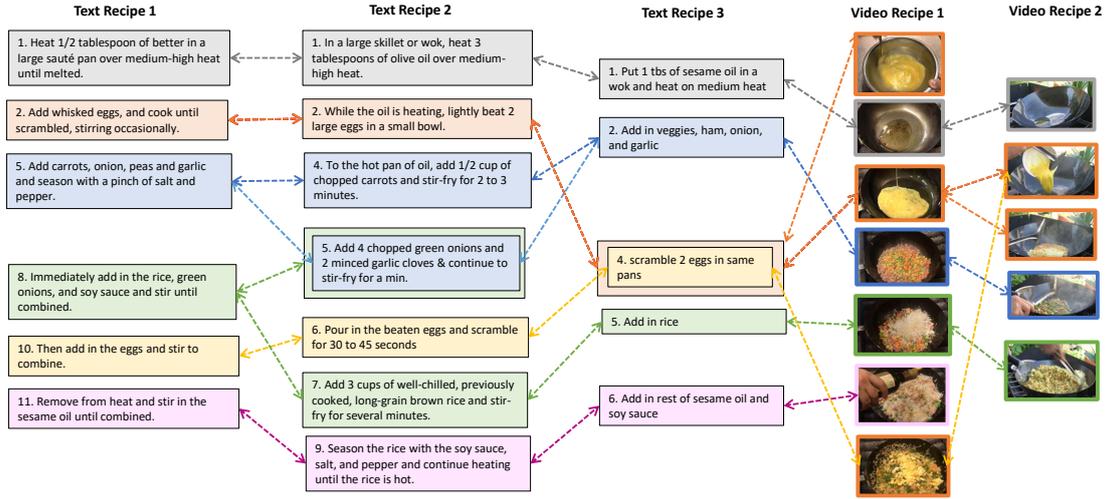}
    \caption{Dish level alignment between three text recipes and two video recipes for \textbf{\textit{fried rice}}. Same colored text boxes (in text recipes) and image borders (in video recipes) indicate instructions that are aligned to each other.}
    \label{fig:multi_text_video_alignment}
\end{figure*}

Although machine learning has seen tremendous recent success in challenging game environments such as Go \cite{schrittwieser2019mastering}, DOTA \cite{berner2019dota}, and StarCraft \cite{starcraft},
we have not seen similar progress toward algorithms that might one day help humans perform everyday tasks like assembling furniture, applying makeup, repairing an electrical problem, or cooking a particular dish.
In part this is because the relevant large-scale multimodal (language, video, audio) datasets are difficult to acquire, even with extensive crowdsourcing \cite{salvador2017learning,sanabria2018how2}.
Unimodal data, though, is abundant on the web (e.g. instructional videos or textual instructions of tasks).
Using language as the link between these modalities, we present an approach for learning large-scale alignment between multimodal procedural data. 
We hope our work, and the resulting released dataset, will help spur research on real-world procedural tasks. 

Recipes in the cooking domain provide procedural instruction sets that are captured -- in large volume -- both in video and text-only forms. Instruction sets in these two modalities overlap sufficiently to allow for an alignment that reveals interestingly different information in the linguistic and visual realms. 
In \autoref{fig:fried_rice_recipe_pair}, for instance, the text recipe (left) and the transcribed video recipe (right) for \textit{shrimp fried rice} vary in word usage, order of instructions and  use of ingredients. Knowing that the highlighted instructions correspond to the same step is useful in understanding potential ingredient substitutions, how the same step can be linguistically described and physically realized in different ways, and how instruction order can be varied without affecting the outcome. 

Motivated by this idea that aligned procedural data can be a powerful source of practical commonsense knowledge, we describe our approach for constructing the \textsc{Microsoft Research Multi-modal Aligned Recipe Corpus}. 
We first extract a large number of text and video recipes from the web.
Our goal is to find joint alignments between multiple text recipes and multiple video recipes for the same dish (see  \autoref{fig:multi_text_video_alignment}). 
The task is challenging, as different recipes vary in their order of instructions and use of ingredients. 
Moreover, video instructions can be noisy, and text and video instructions include different levels of specificity in their descriptions.
Most previous alignment approaches \cite{munteanu-marcu-2005-improving} deal with pairwise alignments. 
Since our goal is to align multiple instruction sets, we introduce a novel two-stage unsupervised algorithm. 
In the first stage, we learn pairwise alignments between two text recipes, two video recipes, and between a text and a video recipe using an unsupervised alignment algorithm (\autoref{sec:pairwise_alignment_algo}). In the second stage, we use the pairwise alignments between all recipes within a dish to construct a graph for each dish and find a maximum spanning tree of this graph to derive joint alignments across multiple recipes (\autoref{sec:joint_alignment_algo}).

We train our unsupervised algorithm on 4,262 dishes consisting of multiple text and video recipes per dish. 
We release the resulting pairwise and joint alignments between multiple recipes within a dish for all 4,262 dishes, along with commonsense information such as textual and visual paraphrases, and single-step to multi-step breakdown (\autoref{sec:data_release}).

We evaluate our pairwise alignment algorithm on two datasets: 1,625 text-video recipe pairs across 90 dishes from the YouCook2 dataset \cite{zhou2018towards}, and a small set of 200 human-aligned text-text recipe pairs across 5 dishes from Common Crawl.
We compare our algorithm to several textual similarity baselines and perform ablations over our trained model (\autoref{sec:experiments}). 
Finally, we discuss how this data release will help with research at the intersection of language, vision, and robotics (\autoref{sec:data_applications}).

\section{Recipe Data Collection}
We describe our approach for collecting large-scale text and video recipes; and  constructing recipe pairs for training our unsupervised alignment algorithm.

\subsection{Common Crawl Text Recipes}
We extract text recipes from Common Crawl,\footnote{https://commoncrawl.org/} one of the largest web sources of text. 
We heuristically filter the extracted recipes\footnote{Details in supplementary.} to obtain a total of 48,852 recipes across 4,262 dishes.
The number of recipes per dish ranges from 3 to 100 (with an average of 6.54 and standard deviation of 7.22).
The average recipe length is 8 instructions. 


\begin{figure}
    \centering
    \includegraphics[scale=0.38]{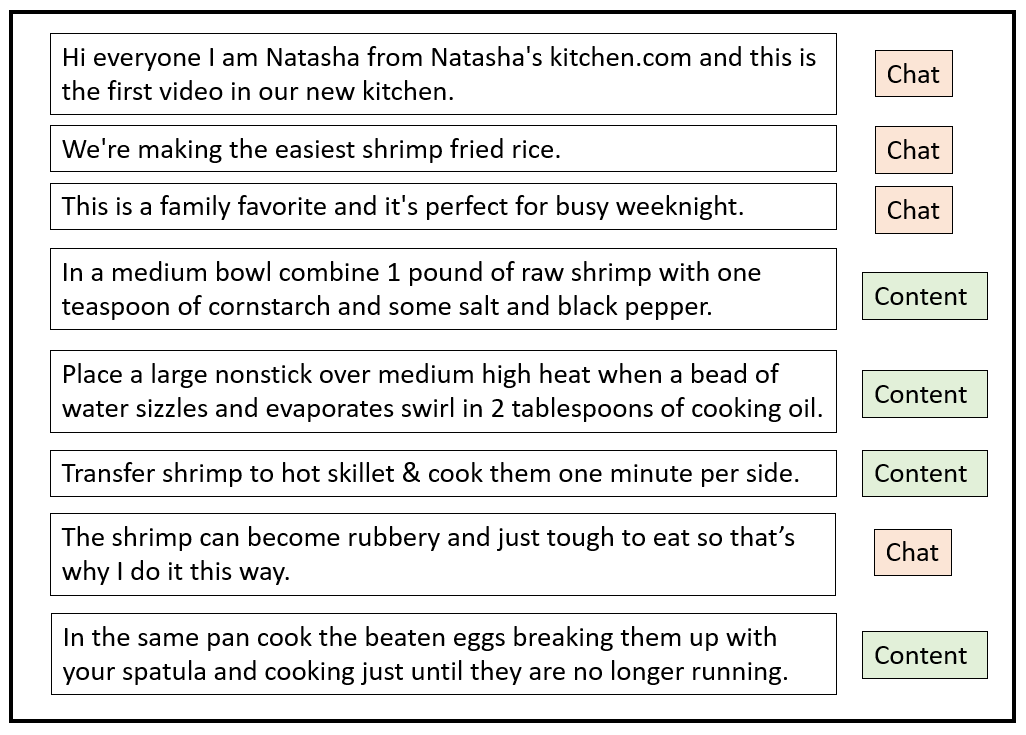}
    \caption{An example transcript of a video recipe with sentences marked as ``chat'' (non-instructional) or ``content'' (instructional).}
    \label{fig:chat_content}
\end{figure}

\subsection{YouTube Video Recipes}
For each dish in the text recipes, we use the dish name with `recipe' appended, e.g. `chocolate chip cookie recipe', as a query on YouTube and extract the top N videos 
where N is proportional to the number of text recipes for that dish\footnote{Details in supplementary.} to obtain a total of 77,550 video recipes.  
We transcribe these videos using the Microsoft Speech-to-Text Cognitive service.\footnote{\url{https://azure.microsoft.com/en-us/services/cognitive-services/speech-to-text/}}

Video recipes, unlike text recipes, contain non-instructional (``chat'') information. For instance, the presenter may give an introduction either of themselves or of the dish at the beginning of the video before diving into the steps of the recipe.  
\autoref{fig:chat_content} contains an example transcript with ``chat'' and ``content'' information marked.
We hypothesize that it is useful to remove such chat information from the transcripts before aligning them to text recipes. 
We build a supervised chat/content classifier using the YouCook2 dataset \cite{zhou2018towards}, an existing instructional cooking video dataset where parts of video that correspond to instructions are annotated by humans. We assume that these parts correspond to content whereas the rest of the video corresponds to chat.\footnote{Details in supplementary.}
We preprocess the transcriptions of all 77,550 videos using this chat/content classifier\footnote{Classifier achieves 85\% F1-score on a held out test set.} to remove all sentences classified as chat. 

\subsection{Recipe Pairs for Training}\label{sec:recipe-pairs-data}
Given N text recipes and M video recipes for a dish, we pair each text recipe with every other text recipe to  get $O(N^2)$ text-text recipe pairs. Similarly, we pair each text recipe with every video recipe to get $O(N*M)$ text-video recipe pairs, and pair each video recipe with every other video recipe to get $O(M^2)$ video recipe pairs.
On closer inspection, we find that some of these pairs describe recipes that are very different from one other, making a reasonable alignment almost impossible. For example, one black bean soup recipe might require the use of a slow cooker, while another describes using a stove. 
We therefore prune these recipe pairs based on the match of ingredients and length\footnote{Details in supplementary.} to finally yield a set of 63,528 text-text recipe pairs, 65,432 text-video recipe pairs and 19,988 video-video recipe pairs.
We split this into training, validation and test split at the dish level. 
Table ~\ref{tab:data_statistics} shows the number of dishes and pairs in each split. 

\begin{table}[t]
    \renewcommand{\arraystretch}{1.3}
    \centering
    \footnotesize
    \begin{tabular}{l|c|c|c}
                        & Train & Val & Test \\
                        \hline
         No. of dishes & 4,065 & 94 & 103 \\
         Text-Text Pairs & 46,054 & 5,822 & 11,652 \\
         Text-Video Pairs & 56,291 & 3,800 & 5,341 \\
         Video-Video Pairs & 19,200 & 274 & 514\\
    \end{tabular}
    \caption{Statistics of our recipe pairs data (\ref{sec:recipe-pairs-data})}
    \label{tab:data_statistics}
\end{table}

\section{Recipe Alignment Algorithm}

We first describe our unsupervised pairwise alignment model trained to learn alignments between text-text, text-video, and video-video recipes pairs. 
We then describe our graph algorithm, which derives joint alignments between multiple text and video recipes given the pairwise alignments.  

\begin{figure*}[t]
    \centering
    \includegraphics[scale=0.58,trim=20 100 0 100]{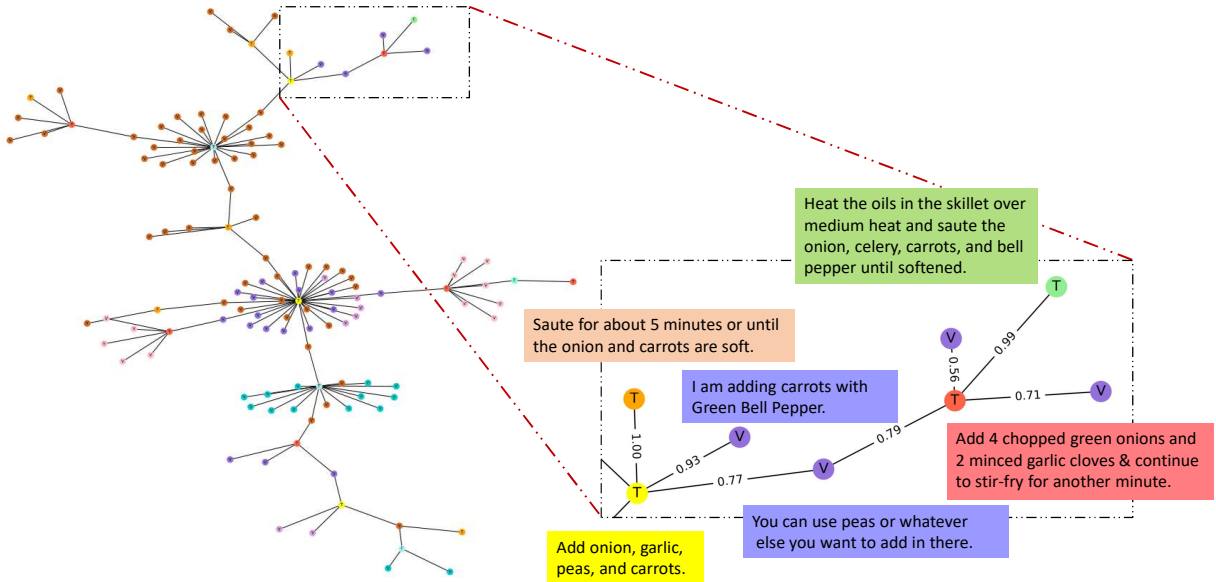}
    \caption{A maximum span tree for \textbf{\textit{fried rice}} dish with text instructions and transcript segments as nodes, alignments as edges, and alignment probabilities as edge weights. Nodes representing text instructions are labeled  ``T". Nodes representing transcript segments are labeled ``V". Each color indicates a different recipe. The bounding box shows a magnified section of the tree with edge weights and the instruction/transcript associated with each node.}
    \label{fig:joint_alignment_graph}
\end{figure*}

\subsection{Pairwise Alignments between Recipes}\label{sec:pairwise_alignment_algo}
Our alignment algorithm is based on prior work \cite{naim2014unsupervised} that learns to align a sequence of natural language instructions to segments of video recording of the same wet lab protocol. 
They first identify the nouns in the text sentences and the blobs (i.e. objects) in video segments.
Given the blobs from $M$ video segments $F = [\textbf{f}^{(1)}, ..., \textbf{f}^{(M)}]$ and the nouns from $N$ sentences $E = [\textbf{e}^{(1)}, ..., \textbf{e}^{(N)}]$, the task is to learn alignments between video segments and text sentences.
They propose a hierarchical generative model which first uses a Hidden Markov Model (HMM) \cite{rabiner1989tutorial,vogel-etal-1996-hmm} to generate each video segment $f^{(m)}$ from one of the text sentences $e^{(n)}$. 
They then use IBM1 model \cite{brown-etal-1993-mathematics} emission probabilities to generate the blobs $\{\textit{f}_1^{(m)},...,\textit{f}_J^{(m)}\}$ in $f^{(m)}$ from the nouns $\{\textit{e}_1^{(n)},...,\textit{e}_I^{(n)}\}$ in $e^{(n)}$ as follows:

\begin{equation}
    P(\textbf{f}^{(m)}| \textbf{e}^{(n)}) = \frac{\epsilon}{(I)^J} \prod_{j=1}^J \sum_{i=1}^J p(\textit{f}_j^{(m)}| \textit{e}_i^{(n)})
\end{equation}

The hidden state in the HMM model corresponds to the alignment between video segment and text sentence, and the state transition probabilities correspond to the jump between adjacent alignments. For computational tractability, a video segment can be aligned to only one sentence (multiple sentences can align to the same video segment)

We use this algorithm to learn pairwise alignments between text-text, text-video and video-video recipes.
Given two recipes (\textit{source} and \textit{target}) of the same dish, we define our alignment task as mapping each text instruction (or video transcript sentence) in the \textit{source} recipe to one or more text instructions (or video transcript sentences) in the \textit{target} recipe.

We make two modifications to the alignment algorithm described above:
First, our recipe pairs, unlike the wet lab protocol data, does not follow the same temporal sequence. The alignment algorithm must thus learn to jump within a longer range. We set the window of jump probabilities at $[-2, 2]$.\footnote{We find that increasing the window beyond 5 decreases performance.} Second, we use transcriptions to learn alignments rather than the objects detected in videos. We hypothesize that the richness of language used in instructional videos may facilitate better alignment with transcripts (as others have observed \cite{malmaud-etal-2015-whats,sener2015unsupervised}).
We use all words (except stop words) in video transcript sentences and all words in text instructions while learning the IBM1 word level probabilities. An instruction in one recipe can be aligned to multiple instructions in the other recipe.

\subsection{Joint Alignment among Multiple Recipes}\label{sec:joint_alignment_algo}

We use the pairwise alignments to derive a joint alignment at the dish level between multiple text and video recipes. 
For each dish, we construct a graph where each node represents an instruction from a text recipe or a transcript sentence from a video recipe. 
We use the pairwise alignments to draw edges between nodes, with alignment probabilities as the edge weights. We include only those edges that have alignment probability greater than 0.5.
The pairwise alignments are directed since they go from the source recipe to the target recipe. We first convert the directed graph into an undirected graph by averaging the edge weights between two nodes and converting directed edges into undirected edges. Note that the resultant graph can have multiple connected components as some recipe pairs may not have any instructions aligned with probability greater than the threshold of 0.5

Our goal is to find a set of jointly-alignable instructions across different recipes. We therefore convert the graph (with cycles) into a forest by running the maximum spanning tree algorithm on the graph.
\autoref{fig:joint_alignment_graph} shows an example tree derived for one of the dishes. 
A path in this tree, that has at most one node from each recipe, constitutes a set of jointly-alignable instructions.
For example, in the magnified section of the tree in \autoref{fig:joint_alignment_graph}, all unique colored nodes in the path from the yellow node to the green node constitute a set of jointly-alignable instructions. 
\section{Experimental Results}\label{sec:experiments}
We describe how we evaluate our pairwise alignment algorithm (from \autoref{sec:pairwise_alignment_algo}). 
We answer the following research questions using our experimentation:
\begin{enumerate}[noitemsep,nolistsep]
\item How does our 
alignment model perform when evaluated on human-aligned recipe pairs?
\item Does our unsupervised alignment model outperform simpler non-learning baselines?
\item How does performance differ when we use only nouns or nouns and verbs instead of all words to learn alignments?
\end{enumerate}

\subsection{Human Aligned Evaluation Set}\label{sec:eval_set}

We evaluate our pairwise alignment algorithm on the following two human annotated datasets:
\paragraph{YouCook2 text-video recipe pairs} The YouCook2 dataset \cite{zhou2018towards} consists of 1,625 cooking videos paired with human-written descriptions for each video segment. These span 90 different dishes. 
We transcribe all videos using the Microsoft Speech-to-Text Cognitive service\footnote{\url{https://azure.microsoft.com/en-us/services/cognitive-services/speech-to-text/}} and separate it into sentences using a sentence tokenizer.
Given a sequence of human-written descriptions and a sequence of transcript sentences, the alignment task is to align each transcript sentence to one of the human-written descriptions. 
We train our pairwise alignment model on the train split of our text-video recipe pairs (from  \autoref{sec:recipe-pairs-data}) and evaluate on the YouCook2 dataset. 
An important difference between the text-video pairs in YouCook2 and in our data is that in YouCook2, the text instructions and the video segments are temporally aligned since the text instructions were specifically written for the videos. In our data, however, the text and the video recipes can differ in order. 

\paragraph{CommonCrawl text-text recipe pairs } 
We randomly choose 200 text-text recipes pairs (spanning 5 dishes) from the test split of our data (\autoref{sec:recipe-pairs-data}) and collect alignment annotations for them using six human experts. 
We show annotators a numbered list of the instructions for the \textit{target} recipe (along with its title and ingredients). We display instructions for the \textit{source} recipe with input boxes besides them and ask annotators to write in the number(s) (i.e labels) of one or more \textit{target} instruction(s) with which it most closely aligns.
Each recipe pair is annotated by three annotators. 
For 65\% of the instructions, two or more annotators agree on a label. For only 42\% of the instructions do all three annotators agree, suggesting that the difficulty level of this annotation task is high. 
We train our pairwise alignment model on the train split of our text-text recipe pairs ( \autoref{sec:recipe-pairs-data}) and evaluate on the 200 human-aligned pairs.

\subsection{Baselines}
Baselines described below align each instruction \footnote{We use the term ``instruction" to mean both text instruction and transcript sentence. } in the \textit{source} recipe to one or more instructions in the \textit{target} recipe. 

\paragraph{Random} We align each instruction in the \textit{source} recipe to a random instruction in the \textit{target} recipe.

\paragraph{Uniform alignment} Given $N$ instructions in the \textit{target} recipe, we divide the instructions in the \textit{source} recipe into $N$ equal chunks and align each instruction in the $i^{th}$ chunk of the \textit{source} recipe to the $i^{th}$ instruction in the \textit{target} recipe.  
For instance, given a \textit{source} recipe $[S1, S2, S3, S4]$ and a \textit{target} recipe $[T1, T2]$, uniform alignment would align $S1$ and $S2$ to $T1$ and $S3$ and $S4$ to $T2$.
More generally, we align the $i^{th}$ instruction in the \textit{source} recipe to the $[(\frac{N}{M}i)^{th}-(\frac{N}{M}(i+1))^{th})$ instruction in the \textit{target} recipe.

\paragraph{BM25 retrieval} We use BM25 \cite{robertson2009probabilistic} as our information retrieval baseline. Given a source and a target recipe pair, we construct a corpus using all instructions in the target recipe. We then use each source instruction as a query to retrieve the top most instruction from the target instruction corpus and align the source instruction to the retrieved target instruction.

\paragraph{Textual similarity}
Given a \textit{source} recipe instruction and a \textit{target} recipe instruction, we define a measure of textual similarity between the two instructions using the following five methods. For each \textit{source} instruction, we compute its similarity score with every \textit{target} instruction and align it to the \textit{target} instruction with the highest score. \\

\begin{table}[t]
    \renewcommand{\arraystretch}{1.1}
    \footnotesize
    \centering
    \begin{tabular}{l|c|c|c}
         \textbf{Methods} & \textbf{Precision} & \textbf{Recall} & \textbf{F1-score}  \\
         \hline
         Random & 18.53 & 14.47 & 14.49 \\
         Uniform alignment & 63.44 & 50.81 & 53.10 \\
         BM25 retrieval  & 48.86 & 39.85 & 38.91 \\
         \hline
         Textual Similarity & & & \\
         \hspace{1.5em} Exact word match & 46.75 & 40.70 & 40.06  \\
         \hspace{1.5em} TF-IDF   & 46.82 & 39.23 & 38.55 \\
         \hspace{1.5em} GloVe & 46.13 & 38.74 & 37.14 \\
         \hspace{1.5em} BERT & 48.83 & 41.48 & 40.89 \\
         \hspace{1.5em} RoBERTa & 50.21 & 42.43 & 42.28 \\
         \hline
         HMM+IBM1 & & &  \\
         \hspace{1.5em}  Nouns & 78.63 & 63.83 & 65.29  \\
         \hspace{1.5em} Nouns+Verbs & 80.56 & 67.90 & 69.00  \\
         \hspace{1.5em} All words & \textbf{81.39} & \textbf{69.27} & \textbf{70.30}  \\
    \end{tabular}
    \caption{Results for text-video recipe alignments on YouCook2 dataset.}
    \label{tab:text-video-results}
\end{table}

\noindent\textit{\textbf{a. Exact word match}}: Given two instructions, we define exact word match as the ratio of the number of common words between the two divided by the number of words in the longer of the two. This gives us a measure of word match that is comparable across instructions of different lengths.\\

\noindent\textit{\textbf{b. TF-IDF}}: We use all the recipes in our training set to create a term frequency (TF)-inverse document frequency (IDF) vectorizer. 
Given an instruction from the evaluation set, we compute the TF-IDF vector for the instruction using this vectorizer.
Given two instructions, we define their TF-IDF similarity as the cosine similarity between their TF-IDF vectors. \\

\noindent\textit{\textbf{c. GloVe}}: We train GloVe embeddings \cite{pennington-etal-2014-glove} on an in-domain corpus of 3 million words put together by combining text recipes and video transcriptions.
Given an instruction, we average the GloVe embeddings \cite{pennington-etal-2014-glove} of nouns and verbs\footnote{We find that using only nouns and verbs outperforms using all words.} 
to obtain its embedding vector.
Given two instructions, we define their embedding similarity as the cosine similarity of their embedding vectors.\\

\noindent\textit{\textbf{d. BERT}}: Given an instruction, we compute its embedding vector using BERT-based sentence embedding \cite{reimers-gurevych-2019-sentence}. We experiment with different variants and find that the BERT-base model trained on AllNLI, then on STS benchmark training set\footnote{\url{https://pypi.org/project/sentence-transformers/}} performed the best for us. 
Given two instructions, we define their BERT similarity as the cosine similarity between their sentence embedding vectors.\\

\noindent \textit{\textbf{e. RoBERTa}}: We also experiment with a variant of the above baseline where we use RoBERTa \cite{liu2019roberta} instead of BERT to compute the sentence embeddings. We use RoBERTa-large trained on AllNLI, then on STS benchmark training set.

\begin{table}[t]
    \renewcommand{\arraystretch}{1.1}
    \footnotesize
    \centering
    \begin{tabular}{l|c|c|c}
    \textbf{Methods} & \textbf{Precision} & \textbf{Recall} & \textbf{F1}  \\
         \hline
         Random & 14.26 & 14.00  & 12.69 \\
         Uniform alignment & 41.38 & 31.85 & 33.22 \\
         BM25 retrieval  & 50.06 & \textbf{55.27} & 49.30 \\
         \hline
         Textual Similarity & & & \\
         \hspace{1.5em} Exact word match & 53.90 & 48.39 & 46.98 \\
         \hspace{1.5em} TF-IDF  & 52.78 & 46.82 & 45.12 \\
         \hspace{1.5em} GloVe & 56.04 & 51.89 &  50.30 \\
         \hspace{1.5em} BERT & 50.72 & 55.07 & 49.10\\
         \hspace{1.5em} RoBERTa & 52.49 & \textbf{55.86} & 50.44 \\
         \hline
         HMM+IBM1 & & &  \\
         \hspace{1.5em} Nouns & 62.11 & 48.99 & 50.73 \\
         \hspace{1.5em} Nouns+Verbs & 64.72 & 50.76 & 52.97 \\
         \hspace{1.5em} All words & \textbf{66.21} & 52.42 & \textbf{54.55}  \\
    \end{tabular}
    \caption{Results for text-text recipe alignment on Common Crawl dataset.}
    \label{tab:text-text-results}
\end{table}

\subsection{Model Ablations}

We experiment with the following ablations of our unsupervised pairwise alignment model (\autoref{sec:pairwise_alignment_algo}):

\paragraph{HMM+IBM1 (nouns)} We use the NLTK\footnote{\url{https://www.nltk.org/}} part-of-speech tagger to identify all the nouns in an   instruction and only use those to learn the IBM1 word-level alignments. This ablation is similar to the model proposed by \citet{naim2014unsupervised} that align objects in videos to nouns in text. 

\paragraph{HMM+IBM1 (nouns and verbs)} We use both nouns and verbs to learn IBM1 word-level alignments. This ablation is similar to the method used in \citet{song2016unsupervised} that align objects and actions in videos to nouns and verbs in text.   

\paragraph{HMM+IBM1 (all words)} We use all words (except stop words) in the \textit{source} and the \textit{target} recipe instructions to learn the word-level alignments.\footnote{Experimental details of HMM+IBM1 model is in supplementary.} 

\subsection{Evaluation Metrics}

Given $M$ source recipe instructions and $N$ target recipe instructions, the alignment task is to label each of the $M$ source instructions with a label from $[0,...,(N-1)]$. 
Given a predicted sequence of labels (from baseline or proposed model) and a reference sequence of labels (from human annotations) for a recipe pair, we calculate the weighted-average\footnote{Calculate metrics for each label, and find their average weighted by the number of true instances for each label.} precision, recall and F1 score. We average these scores across all alignment pairs to compute aggregate scores on the test set.



\subsection{Results}

\paragraph{On text-video alignments} 
\autoref{tab:text-video-results} shows results of our pairwise alignment algorithm compared with baselines on 1,625 human aligned text-video recipe pairs from YouCook2. 
The BM25 baseline outperforms two of the textual similarity baselines. 
Within the textual similarity baselines, RoBERTa outperforms all others suggesting that a pretrained sentence level embedding acts as a good textual similarity method for this alignment task. 
The uniform alignment baseline, interestingly, outperforms all other baselines. This is mainly because in the YouCook2 dataset, the text instructions and the transcript sentences follow the same order, making uniform alignment a strong baseline. 
Our unsupervised HMM+IBM1 alignment model significantly outperforms (with $p<0.001$) all baselines.
Specifically, it gets much higher precision scores compared to all baselines. 
Under ablations of the HMM+IBM1 model, using all words to learn alignments works best. 

\paragraph{On text-text alignments}
\autoref{tab:text-text-results} shows results of our pairwise alignment algorithm compared with baselines on 200 human-aligned text-text recipe pairs from Common Crawl.
Unlike text-video alignments, we find that the uniform alignment baseline does not outperform textual similarity baselines, suggesting that the different re-orderings between text-text recipe pairs makes alignment more challenging. 
Within textual similarity baselines, similar to text-video alignment, RoBERTa outperforms all others.
We believe this is because text recipes tend to share similar vocabulary, making it easier to find similar words between two textual instructions. Video narrators tend to use more colloquial language than the authors of text recipes, making it more difficult to learn alignments using word similarities.
Interestingly, both BM25 and RoBERTa get higher recall than our best HMM+IBM1 model but they lose out on precision. 
This suggests that retrieval models are good for identifying more alignments, albeit with lower precision. 
Our unsupervised HMM+IBM1 model again significantly outperforms ($p<0.001$) all baselines on F1 score.
Under ablations of the HMM+IBM1 model, we again find that using all words to learn alignments performs best. 

\paragraph{Comparing text-video and text-text alignment results}
On comparing \autoref{tab:text-video-results} and \autoref{tab:text-text-results}, we find that textual similarity baselines have overall higher scores on the text-text alignments than the text-video alignments. 
Our HMM+IBM1 model, on the other hand, has overall higher scores on text-video alignments than on text-text alignments. 
We attribute this contrast to the fact that two text recipes have higher vocabulary similarities than a text and a video recipe, resulting in textual similarity baselines to perform well on text-text alignments.
Our HMM+IBM1 unsupervised learning model is able to do better on text-video pairs where the word usage differences are higher. 
Furthermore, the text-video pairs from YouCook2 are temporally aligned whereas the text-text pairs from Common Crawl have several re-orderings making the text-text evaluation set comparatively harder. The supplementary material includes an analysis of alignment outputs.

\section{Data Release}\label{sec:data_release}
We describe the data released in our \textsc{Microsoft Research Multimodal Aligned Recipe Corpus}. 
In all our released data, for text recipes, we include the actual text of the instructions. 
Whereas, for video recipes, we release the URL to the YouTube video with timestamps corresponding to the aligned video segments. 

\begin{figure}[t]
    \centering
    \includegraphics[scale=0.55]{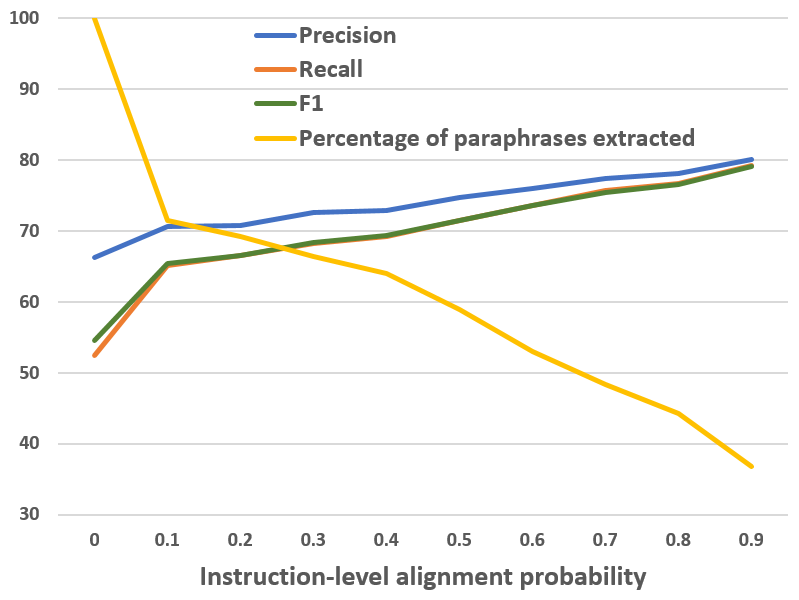}
    \caption{We plot the trade-off between the percentage of paraphrases extracted and the precision, recall and F1 score (as measured by human annotators) with increasing alignment probability threshold on 200 human-aligned text-text recipes pairs.}
    \label{fig:text_text_results_graph}
\end{figure}

\begin{table*}[t]
    \renewcommand{\arraystretch}{1.3}
    \footnotesize
    \centering
    \begin{tabular}{l|l}
       \textbf{Single Step} & \textbf{Multiple Steps}  \\
       \hline
       Beat eggs, oil vanilla and sugar together in a large bowl.  &  1.Beat eggs in large bowl until foamy. \\
       & 2. Add sugar, oil and vanilla mix well.\\
       \hline
       Butter 2 loaf pans and bake 1 hour at 325 degrees. & 1. Pour into greased muffin tins or loaf pans \\
       & 2. Yields about 4 small loaves or 2 large. \\
       & 3. Bake for 25 minutes.\\
       \hline
       Mix the zucchini, sugar, oil, yogurt and egg in a bowl.  & 1. Beat eggs, sugar, oil and vanilla. \\
       & 2. Add zucchini.
    \end{tabular}
    \caption{Three examples of single-step to multi-step breakdown from the pairwise alignments.}
    \label{tab:multi_step_breakdown}
\end{table*}

\subsection{Pairwise and Joint Alignments}
We release the pairwise alignments between recipes of the same dish (derived from \autoref{sec:pairwise_alignment_algo}) for 4,262 dishes. This includes 63,528 alignments between text recipes, 65,432 alignments between text and video recipes; and 19,988 alignments between video recipes.
We also release the joint alignments between multiple text and multiple video recipes within a dish (derived from \autoref{sec:joint_alignment_algo}) for 4,262 dishes.

\subsection{Textual and Visual Paraphrases}
The pairwise alignment algorithm described in \autoref{sec:pairwise_alignment_algo} gives alignment probabilities for each pair of instructions it aligns. We threshold on these alignment probabilities to retrieve textual and visual paraphrases.
Since our goal is to extract large number of high quality paraphrases, we decide on the threshold value by looking at the trade-off between the percentage of paraphrases extracted and their quality as measured by human annotators on 200 human-aligned text-text recipe pairs from our evaluation set (\autoref{sec:eval_set}). 

Figure \ref{fig:text_text_results_graph} shows the trade-off between the precision, recall and F1 score and the percentage of paraphrases extracted with increasing threshold on instruction-level alignment probability.
At $0.5$ threshold, we extract 60\% of the total alignments as paraphrases from our evaluation set.
We use this threshold value of $0.5$ on the pairwise alignments in the training, validation and test sets to extract a total of 358,516 textual paraphrases and 211,703 text-to-video  paraphrases from 4,262 dishes and include it in our corpus. 




\subsection{Single-step to Multi-step breakdown}
The pairwise alignments between text recipes include many instances where one instruction in one recipe is aligned to multiple instructions in another recipe with high alignment probability (greater than $0.9$). \autoref{tab:multi_step_breakdown} shows three such single-step to multi-step breakdown. We extract a total of 5,592 such instances from 1,662 dishes across the training, validation and test sets and include it in our corpus. 




\section{Applications of Our Corpus}\label{sec:data_applications}

We believe that our data release will help advance research at the intersection of language, vision and robotics.
The pairwise alignment between recipes within a dish could be useful in training models that learn to rewrite recipes given ingredient or cooking method based constraints.
The joint alignment over multiple text recipes within a dish should prove useful for learning the types of ingredient substitutions and instruction reordering that come naturally to expert cooks. The textual and visual paraphrases will, we believe, have implications for tasks like textual similarity, image and video captioning, dense video captioning and action recognition. 
The single-step to multi-step breakdown derived from our pairwise alignments may also prove useful for understanding task simplification, an important problem for agents performing complex actions.

Such multimodal data \emph{at scale} is a crucial ingredient for robots to learn-from-demonstrations of procedural tasks in a variety of environments. Collecting such large scale data is prohibitively expensive in robotics since it requires extensive instrumentation of many different environments. Other example applications are learning to ground natural language to physical objects in the environment, and catching when humans are about to commit critical errors in a complicated task and offering to help with corrective instructions.
 
\section{Related Work}

\noindent\paragraph{Alignment Algorithms}
Our unsupervised alignment algorithm is based on \citet{naim2014unsupervised}, who propose a hierarchical alignment model using nouns and objects to align text instructions to videos. 
\citet{song2016unsupervised} further build on this work to make use of action codewords and verbs. 
\citet{bojanowski2015weakly} view the alignment task as a temporal assignment problem and solve it using an efficient conditional gradient algorithm. 
\citet{malmaud-etal-2015-whats} use an HMM-based method to align recipe instructions to cooking video transcriptions that follow the same order.
Our work contrasts with these works in two ways: we learn alignments between instructions that do not necessarily follow the same order; and our algorithm is trained on a much larger scale dataset. 



\noindent\paragraph{Multi-modal Instructional Datasets}
\citet{marin2019learning} introduce a corpus of 1 million cooking recipes paired with 13 million food images for the task of retrieving a recipe given an image. 
YouCook2 dataset \cite{zhou2018towards} consists of 2,000 recipe videos with human written descriptions for each video segment. 
The How2 dataset \cite{sanabria2018how2} consists of 79,114 instructional videos with English subtitles and crowdsourced Portuguese translations.
The COIN dataset \cite{tang2019coin} consists of 11,827 videos of 180 tasks in 12 daily life domains. 
YouMakeup \cite{wang2019youmakeup} consists of 2,800 YouTube videos, annotated with natural language descriptions for instructional steps, grounded in temporal video range and spatial facial areas.

\noindent\paragraph{Leveraging Document Level Alignments} 
Our work relies on the assumption that text recipes and instructional cooking videos of the same dish are comparable.
This idea has been used to extract parallel sentences from comparable corpora to increase the number of training examples for machine translation \cite{munteanu-marcu-2005-improving,abdul-rauf-schwenk-2009-use,smith-etal-2010-extracting,gregoire-langlais-2018-extracting}. 
Likewise, TalkSumm \cite{lev-etal-2019-talksumm} use the transcripts of scientific conference talks 
to automatically extract summaries. 
\citet{zhu2015aligning} use books and movie adaptations of the books to extract descriptive explanations of movie scenes.

\noindent\paragraph{Related Tasks}
A related task is localizing and classifying steps in instructional videos \cite{alayrac2016unsupervised, zhukov2019cross} where they detect when an action is performed in the video whereas we focus on describing actions.
Dense event captioning of instructional videos \cite{zhou2018end,li2018jointly,hessel-etal-2019-case} relies on human curated, densely labeled datasets whereas we extract descriptions of videos automatically through our alignments.




\section{Conclusion}\label{sec:conclusion}

We introduce a novel two-stage unsupervised algorithm for aligning multiple text and multiple video recipes. We use an existing algorithm to first learn pairwise alignments and then use a graph-based algorithm to derive the joint alignments across multiple recipes describing the same dish. 
We release a large-scale dataset constructed using this algorithm consisting of joint alignments between multiple text and video recipes along with useful commonsense information such as textual and visual paraphrases; and single-step to multi-step breakdown.

Although our dataset focuses on the cooking domain, our framework should generalize to any domain with abundant volumes of unstructured-but-alignable multi-modal data. DIY (Do-It-Yourself) videos and websites, for instance, are an obvious next target. We also envision extending this work by including audio and video features to enhance the quality of our alignment algorithm. 
Ultimately, we believe this work will further the goal of building agents that can work with human collaborators to carry out complex tasks in the real world.

\section*{Acknowledgments}
We would like to thank Harpreet Sawhney, Roshan Rao, Prasoon Goyal, Dilip Arumugam and Raymond J. Mooney for all their help.
We would also like to thank the four anonymous reviewers for their useful comments and suggestions. 

\bibliography{anthology,main}

\bibliographystyle{acl_natbib}

\appendix

\section{Supplemental Material}
\label{sec:supplemental}

In this supplementary, we describe the details of our data collection process (\autoref{sec:data_collection}), experimental details of our algorithm (\autoref{sec:experimental_details}) and provide analysis of our alignment outputs (\autoref{sec:error_analysis}). 

\subsection{Details of Data Collection}\label{sec:data_collection}

\subsubsection{Common Crawl Text Recipes}
We use recipe data from Common Crawl \footnote{https://commoncrawl.org/} that has metadata formatted according to the Schema.org Recipe schema \footnote{https://schema.org/Recipe} including title, ingredients, instructions, and a URL to the recipe source. 
There were originally 3.2 million recipes extracted from Common Crawl. We filter the data by limiting the data to recipes with instructions written in English, removing recipes with titles that are longer than 5 words, removing duplicate recipes, removing recipes where the recipe title contains words that are not in the top 50\% most common words that occur in the recipe titles, and removing recipes with fewer than 2 steps. After filtering the data, we clustered the recipes into dishes using exact match on the recipe titles. We only retain recipes from dishes that have at least three recipes. 
The final dataset has a total of 4,262 dishes and 48,852 recipes with an average of 8 instructions per recipe. 

\subsubsection{YouTube Video Recipes}
Given the dish names from the text recipes, we extract YouTube video recipes for each of the dishes. The number of videos extracted for each dish is proportional to the number of text recipes found for that dish. For instance, for a more popular dish like \textit{chocolate chip cookies}, we would extract more text and video recipes than for a less popular dish like \textit{creme brulee}.
The number of videos extracted ranges from 3 to 100.

\subsubsection{Chat/Content Classifier}

Instructional cooking videos can contain a lot of non-instructional content (``chat''). For example, the person cooking the dish often introduces themselves (or their video channel) at the beginning of the video. They sometimes also introduce the dish they are going to prepare and suggest pairings for the dish. The non-instruction content are often found in the beginning and towards the end of the video but there are several instances of ``chat'' interspersed with instructional content as well.
Since we wish to align these videos to text recipe instructions that do not contain non-instructional information, we need a way to remove non-instructional content.
We train a supervised neural network based classifier for this task.

We train our classifier using the YouCook2 dataset \cite{zhou2018towards} of 1,500 videos across 90 dishes. This dataset was created by asking humans to identify segments of a video that correspond to an instruction and annotate each segment with an imperative statement describing the action being executed in the video segment.
We make the assumption that the transcript sentences that are included within an annotated video segment are instructional whereas those that are not included within an annotated video segment are non-instructional. 
We first transcribe all 1,500 videos in the dataset using a commercial transcription web service. 
We split the transcription into sentences using a sentence tokenizer. We label a transcript sentence with the label $1$ if the corresponding video segment was annotated and with the label $0$ if it was not. 
We get a total of 90,927 labelled transcript sentences which we split by dishes into the training (73,728 examples), validation (7,767 examples) and test (9,432 examples) sets. 

We use an LSTM (long-short term memory) model \cite{hochreiter1997long} with attention \cite{luong-etal-2015-effective} to train a binary classifier on this data. We initialize (and freeze) our 300-dimensional word embeddings using GloVe \cite{pennington-etal-2014-glove} vectors trained on 330 million tokens that we obtain by combining all text recipes and transcript sentences. 
We use the validation set to tune hyperparametrs of our LSTM classifier (hidden size: 64, learning rate: 0.00001, batch size: 64, number of layers: 1). 
Our chat/content classifier achieves $86.76$ precision, $84.26$ recall and $85.01$ F1 score on the held out test set. 


\subsubsection{Recipe Pair Pruning Strategy}

We define the following two pruning strategies to reduce the number of extracted recipe pairs:

\textbf{Ingredient match}: 
Each of our text recipes from Common Crawl contains an ingredients list.
Video recipes from YouTube however do not contain ingredient lists. We therefore estimate the ingredients for video recipes using text recipes of the same dish. 
We construct a set of ingredients at the dish level by combining all ingredients of the text recipes within that dish.
We then use this dish-level ingredients information to identify ingredient words from the words of video transcriptions. 
Given a recipe pair, we compare the ingredients of the two recipes and if the percentage of ingredients that match is below a threshold, we remove the pair. For text-text and text-video recipe pair, we set this threshold to be 70\%, whereas for video-video recipe pair, we set this threshold to be 90\% (since video-video recipe pairs tend to be more noisy).

\textbf{Instruction length match}: 
For text-text recipe pairs, if number of instructions in one recipe is more than double the number of instructions in another recipe, we remove the pair.
For video recipes, if there are more than 100 sentences in the transcript after removing the background sentences, we remove that video recipe.

\subsection{Details of HMM+IBM1 Model}\label{sec:experimental_details}

We train the HMM+IBM1 pairwise alignment model on three kinds of recipe pairs: text-text, text-video and video-video. 
The lower level IBM1 model works on words of text instruction or transcript sentences. 
The vocabulary size of all the text recipes from 4,262 dishes put together totals to 48,609 words. Since most words do not appear very frequently across the text recipes corpus, we reduce the vocabulary size to 13,061 by removing words that occur fewer than 5 times in the training set.
Likewise, we reduce the vocabulary size of video recipe transcriptions to 16,733 words (from 88,744 words) by removing words that occur fewer than 15 times in the training set. 
We first train the HMM+IBM1 model for 3 iterations with a jump range of $[-1,0,+1]$ and further train it for 2 iteration with a jump range of $[-2,0,+2]$. We find that warm starting the model with a shorter range helps the model to learn better alignments. 

\subsection{Alignment Output Analysis}\label{sec:error_analysis}

\autoref{tab:text_text_alignment} shows the alignment between two text recipes for \textit{chocolate chip cookies} obtained by our pairwise algorithm. The alignment task here is to align each instruction in the source recipe to one of the instructions in the target recipe. The table displays all the instructions in the source recipe in the second column. The first column of the table displays instructions from the target recipe that aligns to the source recipe instruction in the same row. The sentence level probabilities are shown in the last column.

We can see the reordering between the two recipes by comparing the instruction indices. We see that instructions 0 to 2 from the source are aligned to target instructions with very high probabilities suggesting they are close paraphrases. Instruction 3 and 8 from the source, on the other hand, are aligned with comparatively lower probabilities to the target and we can see that in these two cases, the two instructions do differ in meaning. Instructions 6,7 and 8 (in source) aligned to instruction 11 (in target) is an example of single step to multi-step breakdown.

\begin{table*}[t]
    \renewcommand{\arraystretch}{1.2}
    \centering
    \footnotesize
    \begin{tabular}{l|l|l}
   \textbf{ Target recipe instruction} & \textbf{Source recipe instruction} & \textbf{Probability}\\
    \hline
    0: Preheat your oven to 350 degrees F. &  0: Preheat the oven to 350 degrees F.& 0.9999 \\
    \hline
    2: In the bowl of your mixer cream &
    1: In a large bowl or the bowl of a stand & 0.9998 \\
    together your butter and sugars until 
    & mixer cream the butter sugar brown sugar  & \\
     light and fluffy about 3-5 minutes. & 
     eggs \& vanilla together until smooth \& fluffy. & \\
     \hline
     1: Sift together the flour baking soda &
     2: In another bowl whisk together & 0.9997\\
     baking powder and salt into a medium & 
     the flour salt baking powder and baking soda. & \\
     sized bowl and set aside. &  & \\
     \hline
     4: Add in the vanilla and mix. & 
     3: Add this to the butter mixture & 0.6889 \\
     & and mix until well combined. & \\
     \hline
     6: Fold in your chocolate until evenly &
     4: Stir in the chocolate chips. & 0.9820 \\
     added throughout the dough. & & \\
     \hline
     8: Scoop your dough out onto the sheets. &
     5: Form the dough into golf-ball sized  & 0.9997\\
     & balls and place them about 2 inches & \\
     & apart on a baking sheet. & \\
     \hline
     11: Bake 10-12 minutes for smaller cookies  &
     6: Bake for 9-10 minutes just until the & 0.9912\\
     or 18-20 minutes for larger cookies. & 
     edges start to brown lightly. & \\
     \hline
     11: Bake 10-12 minutes for smaller cookies  &
     7: Do not overbake them or they will be & 0.9528\\
     or 18-20 minutes for larger cookies. & 
     crispy rather than chewy. & \\
     \hline
     11: Bake 10-12 minutes for smaller cookies &
     8: They still look underbaked when you & 0.6465\\
     or 18-20 minutes for larger cookies. & 
      take them out but will firm up as they cool. & \\
     \hline
     12: Allow the cookies to cool slightly &
     9: Let them cool on the pan for about 5 & 0.9973\\
     on your baking sheet then move them to &
     minutes and them move to a wire rack & \\
     another surface to cool completely. &
      to cool completely. & \\
      \hline
     14: Store in an air-tight container at & 
     10: Cookies will keep for 7 days in & 0.8309 \\
     room temperature for up to 3 days or & 
     a sealed container at room temperature. & \\
     freeze for up to 2 months. & & \\
    \end{tabular}
    \caption{Alignment between two text recipes of \textit{chocolate chip cookie} with their sentence level probabilities. }
    \label{tab:text_text_alignment}
\end{table*}

\end{document}